\documentclass[times, 3p]{elsarticle}
\usepackage{makecell}
\usepackage[utf8]{inputenc} % allow utf-8 input
\usepackage[T1]{fontenc}    % use 8-bit T1 fonts
\usepackage{hyperref}       % hyperlinks
\usepackage{url}            % simple URL typesetting
\usepackage{booktabs}       % professional-quality tables
\usepackage{amsfonts}       % blackboard math symbols
\usepackage{nicefrac}       % compact symbols for 1/2, etc.
\usepackage{microtype}      % microtypography
\usepackage{lipsum}
\usepackage{graphicx} 
 \usepackage{caption}
\usepackage{subcaption}
\usepackage[linesnumbered,boxed]{algorithm2e}
\usepackage{amsthm,amsmath,amssymb,mathrsfs}
\usepackage[table]{xcolor}
\usepackage{multirow}
\newcommand\sbullet[1][.5]{\mathbin{\vcenter{\hbox{\scalebox{#1}{$\bullet$}}}}}
 
\DeclareMathOperator*{\argmax}{argmax} % thin space, limits underneath in displays
\DeclareMathOperator*{\sign}{sign}

\begin{document}
\begin{frontmatter}
\title{Deep-Attack over the Deep Reinforcement Learning}

\author[NPU]{Yang Li\corref{mycorrespondingauthor}}
\ead{liyangnpu@nwpu.edu.cn}
\author[NPU]{Quan Pan}
\ead{quanpan@nwpu.edu.cn}
\author[NTU]{Erik Cambria}

\cortext[mycorrespondingauthor]{Corresponding author}
\ead{liyangnpu@nwpu.edu.cn}

\address[NPU]{Northwestern Polytechnical University, China}
\address[NTU]{Nanyang Technological University, Singapore}

\begin{abstract}
% Recent adversarial attack development has made reinforcement learning more vulnerable. There are different ways to deploy attacks on it. The key is how to choose the right time to attack. 
Recent adversarial attack developments have made reinforcement learning more vulnerable, and different approaches exist to deploy attacks against it, where the key is how to choose the right timing of the attack.
% Some works attempt to design an attack evaluation function to select the key point, and if the value is greater than a certain threshold, the attack will be carried out.
Some work tries to design an attack evaluation function to select critical points that will be attacked if the value is greater than a certain threshold.
This approach makes it difficult to find the right place to deploy an attack without considering the long-term impact. In addition, there is a lack of appropriate indicators of assessment during attacks. 
To make the attacks more intelligent as well as to remedy the existing problems, we propose the reinforcement learning-based attacking framework by considering the effectiveness and stealthy spontaneously, while we also propose a new metric to evaluate the performance of the attack model in these two aspects.
% at the meantime, we also proposed a new metric to evaluate these two aspects.
Experimental results show the effectiveness of our proposed model and the goodness of our proposed evaluation metric. Furthermore, we validate the transferability of the model, and also its robustness under the adversarial training. 
\end{abstract}
\end{frontmatter}
% \begin{IEEEkeywords}
% Adversarial Attack, Deep Reinforcement Learning, Adversarial Training.
% \end{IEEEkeywords}

\section{Introduction}
\label{sec:intro}
With help of deep learning~\cite{li2021graph}, deep reinforcement learning (DRL)~ has developed rapidly in the past few years and has become an important algorithm in many applications, e.g., automatic navigation~\cite{li2019deep}, text generation~\cite{li2018generative}, sentiment analysis~\cite{peng2021phonetic}, etc.   
% It is to train an advanced policy network and guide agents to take the optimal action based on the environment state. 
In a deep reinforcement learning, it is to train an advanced policy network and guide the agent to take the optimal action in accordance with the state of the environment~\cite{zhang2021real}.
A policy network is usually a deep learning model to approximate the action-value function.
However, the vulnerability of the deep learning model which is discovered by Szegedy et al.~\cite{szegedy2014intriguing} has made DRL unstable when tackling different types of adversarial attacks. 
Attacks are usually deployed by adding some carefully crafted perturbations to the input and guiding the deep learning model to give incorrect outputs with high confidence.
Unlike a typical supervised/unsupervised learning task which can be interpreted as a two-element tuple, i.e., $<\mathscr{X}, \mathscr{Y}>$ (where $\mathscr{X}$ indicates the data and $ \mathscr{Y}$ indicates the label), there are at least four elements i.e., $<S, A, \pi, R>$  in a typical DRL (where $A$ is the action set, $S$ is the state set, $R$ is the reward set, and $\pi$ is the policy network.). 
Therefore, DRL would be theoretically more vulnerable than supervised/unsupervised learning.
And in fact, there have been many studies on attacking DRL by deploying perturbations on different variables.
In general, depending on the attack variables, they can be classified as reward-based attacks, observation-based attacks, action-based attacks, and policy-based attacks~\cite{chen2021survey}, and all of these attack strategies are designed to mislead the agent into taking the wrong action.
% As it is not a one-time attack, an attack in DRL is more difficult than a general domain as it is a series of steps when executing actions with a fully trained policy network.
Because it is not a one-time attack, an attack in DRL is more difficult than an attack in the general domain due to the series of steps involved in executing an action with a well-trained policy network.
In addition, there is no clear definition of what constitutes a successful attack. Instead, there are two main constraints for a successful attack on an image, namely effectiveness and stealth. 
Therefore, it is necessary to propose evaluation metrics for the effectiveness of attacks on DRLs.
% Specifically, there are two constraints for a successful attack which are attacking effectiveness and attacking stealthily.

% However, one important problem is to decide when to attack (i.e., when the attack is expected to be effective) and how to attack (i.e., selecting the amount of noise to add),  
% Among these attacks, the adversary also has the option to choose not to attack, so it has to determine when to attack and how to attack, and both in when and how the long-term effects of the attacks must be taken into account. 
% In particular, it only inputs adversarial states and observes the consequences of the corresponding victim’s model output actions to learn an attack policy. 
% In addition, unlike most existing black-box attacks, it also takes into account the sequential nature of the problem, and the need to optimize both long-term performance and cost.

Attacking effectiveness means an attack confusing the model successfully with few training samples, however in DRL we need to consider both long-term success and short-term success as it is a series of actions.
% however, in DRL, we need to take long-term success and short-term success into consideration at the same time as it is a series of actions. 
% Therefore, we can categorize them into are generally two types of attacking success, namely short-term success and long-term success. 
Specifically, short-term success means an attack affects the agent instantly and misleads it taking the wrong action immediately. This type of attack focuses on immediate effectiveness.
While long-term success means an attack destroys the task with the fewest actions. 
This type of attack has been widely applied in many attacking algorithms~\cite{tretschk2018sequential,chen2019adversarial}.
The stealthiness of an attack is another important feature, which is important in practical applications because if an attack can be easily detected, it will simply be ended with brute force. 
A perfect attack is like the advanced persistent threat (APT) attack in network security, it will infiltrate the target for a long time until the final attack is carried out to destroy the mission.  

However, there are seldom works combining the two attacks together. 
Generally, if we can find the key points for attacking which may have a long-term effect, the attack will be more effective. 
But the truth is that it is difficult to find them as they are not connected with future action directly.
However, as we mentioned above, the short-term attack is an instant action, while the long-term attack is global action.  
If we can find the connection between these two types of attacks, their effectiveness will improve a lot.
In this paper, to cope with such a problem, a general attacking framework (i.e., attacker) with Deep Q-Network (DQN) is proposed, the goal of the attacker policy network is to learn the timing of the attack. 
Generally, it is to find a policy $\pi^{*}$ that maximizes the total rewards from all perturbed states, which is illustrated as follows:
\begin{equation}
\label{equ:1}
\pi^{*} = \mathop{\arg\!\max}_{r} \{E[\sum_{t=0}^{M_{X}}\gamma r^{X}(\pi(ATK(s_{t});\theta^{X}))]\}
\end{equation}
where $M_{X}$ is the total step and $r^{X}(\cdot)$ denotes the attacking rewards function, $\pi(\cdot)$ usually is a neural network which can be parameterized with $\theta^{X}$, $ATK(\cdot)$ is the attacking method, and $\gamma$ is the discount factor.  
To ensure the stealthy and effectiveness of attacks, we designed the short-term reward $r_{str}$ and long-term reward $r_{ltr}$ to guide the agent, at the meantime, these two rewards can also be appropriate evaluation indicators.
The contribution of this paper can be summarized as follows:
\begin{itemize}
\item Proposed a novel attacking framework which not only considers the long-term attack but also focuses on the short-term attack based on deep reinforcement learning;
\item Proposed a new evaluation metric for the attacking effectiveness and stealth;
\item Validated the effectiveness of the proposed model with empirical experiments.
\end{itemize}

This paper is organized as follows: Section~\ref{sec:related_works} describes the related works about the adversarial attack in reinforcement learning. Section~\ref{sec:preliminary} provides the background of DQN and the traditional attacking method about critical point selection. Section~\ref{sec:methods} specifies how we design our attack method and Section~\ref{sec:experiments} summarizes the evaluation results. Finally, we conclude our work and give future works in Section~\ref{sec:conclusions}.

% We need to give a clear definition of damage and attack cost. And also need to make a clear definition of what is good damage or attack. A good attack should be stealthy and strong when it is needed. But sometimes it needs stealth and weak.

\section{Related Works}
\label{sec:related_works}
There are increasing researches on the adversarial attack and defense that exists in reinforcement learning, we will give a brief summarizing of existing works in terms of adversarial attack and adversarial defense separately.

\subsection{Adversarial Attack}
There are plenty of works about the adversarial attack in DRL~\cite{wu2021adversarial}, and based on the attacking variables, they can be classified into the observation-based attack, reward-based attack, action-based attack,  policy-based attack, and environment-based attack~\cite{chen2021survey}. 
Observation-based attack means the attack occurs on the state observed by the agent, and the attack usually is deployed with the image-based attacking method. 
However, it is difficult to distinguish the difference between the observation- and environment-based attack as the state and environment can be the subset of the observation in general. 
For example, Huang et al.~\cite{huang2017adversarial} who was the first one to explore the vulnerability in reinforcement learning deployed the attacks (i.e., Fast Gradient Sign Method (FGSM)) over the state directly. 
Behzadan et al.~\cite{behzadan2017vulnerability} deployed the attack over the state to induce the agent with a transfer learning-based method on the DQN.
To find the right time of attacking,  Lin et al.~\cite{lin2017tactics} proposed the Strategically-Timed Attack which only selects the key framework to conduct the attack by designing a judgment function based on the state-action pair. The attacking method they deployed is Carlini and Wagner attack (C\&W)~\cite{carlini2017magnet}.
Similarly,  Kos et al.~\cite{kos2017delving} adopted value function to guide the attack by judging whether the value is higher than a certain threshold,  and Russo et al.~\cite{russo2019optimal} applied the Markov decision process to select the attacking frame to poison. 
All those attacks mentioned above need to know the parameters of the model, i.e., white-box attack.
While there also are works on the black-box attack that can access only the input and output of the RL. For example, Huang et al.~\cite{huang2017adversarial} utilized a surrogate policy network with a rollout technique to construct the adversarial state, and experimental results showed the usefulness of attack.
Zhao et al.~\cite{zhao2020blackbox} proposed an approximation model to deploy the black-box attack by predicting the target agent's future actions based on their time-series information.
And similar works can be found in works~\cite{behzadan2017vulnerability,xiao2019characterizing}.

Apart from manipulating the environment or state that agents are in, some works try the decrease the performance by crafting the rewards, which is called reward-based attack listed in work~\cite{chen2021survey}.
For example, Tretschk et al.~\cite{tretschk2018sequential} adopted the adversarial transformer network to craft the adversarial rewards over the victim agent with a sequence of attacks, and the agent is misled in the opposite reward direction over time.
While Han et al.~\cite{han2018reinforcement} applied a more direct way to conduct the reward-based attack by flipping a certain number of rewards signs. 
Kiourti et al.~\cite{kiourti2019trojdrl} utilized the Trojan technique to attack the target policy network by deploying the in-band reward modification.  
Perturb over the action space is another approach to attack the agent, for example, 
Lee et al.~\cite{lee2020spatiotemporally} conducted the action space attack by reformulating the optimization problems that distribute the attacks across the action and temporal dimensions. 
In this paper, the goal is to learn an intelligent attacker which can deploy the attack at the right place by considering the effectiveness and stealthy spontaneously. Therefore, the environment is where we would like to deploy the attack.

% \subsection{Adversarial Attack over Neural Network}

% A black-box threat model to identify the sequence of attacks to maximize the damage on the victim, and to minimize the attacks’ cost, and where it is assumed the attacker to have minimal information of the target.

% Generally, attacks may happen in the training stage, and attacks that happen in the testing/deploying stage. In both cases, we can classify those attacks into white-box attacks and black-box attacks:
% White-box attacks: In white-box attacks, the adversaries have total knowledge about the target model, including algorithm train, data distribution, model parameters. Given a benign input, the attacker can compute adversarial examples as an optimization algorithm. The accessing of the model’s internal data for white-box attacks corresponds to a strong adversarial capability, but such an amount of knowledge of the victim’s model is unrealistic in real-world applications.
% Black-box attacks: they assume a more realistic threat model, where the adversaries do not know about target mode (Sun et al., 2018). The adversary is restricted from interacting via queries and classification outputs. It uses this information about input/output pairs to analyze the model’s vulnerability.
\subsection{Adversarial Defense}

All defense methods applied in reinforcement learning can be categorized into those three types, i.e., adversarial training, robust learning, and adversarial detection. 
Adversarial training is to add the adversarial samples into the training data, the main goal is to improve the generalization of the model. Since Goodfellow et al.~\cite{goodfellow2014explaining} proposed FGSM to conduct the adversarial training over the image classification, this method has been successfully applied in reinforcement learning.
For example,  Kos et al.~\cite{kos2017delving} and Pattanaik et al.~\cite{pattanaik2018robust} applied the FGSM to obtain adversarial examples constantly which helps the classification model acquire the resistance to adversarial examples.  
While the limitation of this method is that it cannot defend the outlier that does not exist in the training set.
Apart from adding the adversarial samples constantly, some researchers do it differently,  for example, Behzadan et al.~\cite{behzadan2017whatever} validated that if the adversarial samples generated by FGSM are added to the training set with a certain probability (i.e., $P$), better defense effects can also be achieved. And empirical results of DQN show that when setting $P$ to 0.2 or 0.4, DQN can obtain a normal performance after the adversarial training.
Based on this work, Behzadan et al.~\cite{behzadan2019analysis} adopted the $\varepsilon $-greedy search to find a more stable and efficient way to add adversarial samples in their most recent work.
Another adversarial training approach is through the min-max game, For example, Gu et al.~\cite{gu2019adversary} formulated the task as a zero-sum dual-agent Markov game by splitting the agent into adversarial and protagonist, and the robustness is achieved from the min-max game between them.
Similarly, Pinto et al.~\cite{pinto2017robust} introduced two agents to play the zero-sum discount game to ensure the robustness of the policy learning. 
Different from the adversarial playing, Behzadan et al.~\cite{behzadan2018mitigation} adopt an equivalent model as the noisy network to generate the adversarial samples via FGSM.
Neklyudov et al.~\cite{neklyudov2018variance} applied the Gaussian variance layer to generate the adversarial samples, and empirical results show that this method is effective in improving the ability of exploration and robustness of agents.
In this paper, we mainly focus on adversarial training and explore how adversarial training can enhance the robustness of DRL.

\section{Preliminary}
\label{sec:preliminary}
\subsection{Deep Q-learning Network}
Deep Q-learning is a classical reinforcement learning method, and it can be formulated as a Markov Decision Process (MDP), which is defined with a tuple $<S, A, \pi(\theta), R, \gamma, \mu_{0}>$ where $S$ is the state space, $A$ is the action space, $\pi(\theta)$ is the policy network, $R$ is the reward function, and $\gamma$ denotes discount factor indicating how much the immediate reward will affect the future reward and $\mu_{0}$ is the initial state. 
The agent tries to learn a policy network to interact with the environment efficiently. And the DQN aims to find an optimal policy network by solving the Bellman optimal equation:
\begin{equation}
Q(s,a)=E[r+\gamma \max_{a'}Q^{*}(s',a')]
\end{equation}
Where $r$ denotes the reward of the agent, $s'$ denotes the current state, and $a'$ is the action space.
The general way is to train the network by minimizing the temporal different loss (e.g., squared Bellman error). 
And the best action is selected with the maximum Q value.

\subsection{Strategically Timed Selection}
As it is the sequential task for the agent to interact with the environment, to reduce the attacking frequency, and increase the attacking effectiveness, lots of studies focus on how to select the critical point (frame) to deploy the attack. 
A more general way is to select by inner driving where the attack is only deployed when the agent strongly prefers a specific action. And it is usually decided by the predefined attack score.
% which is decided by an attack score. 
In the policy gradient-based method, the attack score  is usually defined as in Equation~\ref{equ:thresh}.
\begin{equation}
\label{equ:thresh}
c(s_{t}) =  \max_{a_t}\pi(s_{t}, a_{t}) - \min_{a_{t}}\pi (s_{t},a_{t})
\end{equation}
where $\pi$ is the policy network which maps state-action pairs to a probability in action space,
same intuition is always deployed in value-based methods (e.g., DQN), where the attack score is defined in Equation~\ref{equ:thres1}.
\begin{equation}
\label{equ:thres1}
c(s_{t}) = \max_{a_{t}}\Phi(Q(s_{t},a_{t})) - \min_{a_{t}}\Phi(Q(s_{t},a_{t}))
\end{equation}
where $Q$ is the value network, which  $\Phi$ is the score function (e.g., softmax function). 
The attack is conducted when the attack score is larger than a threshold, which helps increase the attack effectiveness. 
While in some papers, to make the attack stealthy, they deployed the attack only when the variance of $Q$ value (in Equation~\ref{equ:variance}) is low.  

\begin{equation}
\label{equ:variance}
v(s_{t}) = \frac{1}{|A|-1} \sum_{i=1}^{|A|}(Q(s_{t},a_{i}) - \frac{1}{|A|}\sum_{j=1}^{|A|}Q(s_{t},a_{j}))
\end{equation}
Therefore, the attack rules are quite opposite when considering the different requirements.
However, this type of attack depends on the threshold that is decided manually, and also they are lacking the comprehensive evaluation metric about the attacking effectiveness and stealthy. 

\subsection{Attacking Methods}
In this paper, the attack is deployed in a white-box environment where we know the parameters of the model, i.e., we need to know the actions of the target agent, as well as information about the state of each frame after the agent performs the action.
Generally, for white-box attacks, the adversarial examples can be learned from as $x_{adv} = \underset{{\|x'-x\|_{\infty}\leq \varepsilon}}{\argmax} l(h(x'),y)$, where $x'$ denotes the crafted adversarial sample, and $l(\cdot)$ represents the loss function between the output and the correct label.
There are several efficient attacking approaches ($ATK$ for short) to craft the adversarial examples method, e.g., FGSM~\cite{goodfellow2014explaining}, Projected Gradient Descent (PGD)~\cite{madry2018towards}, and C\&W~\cite{carlini2017towards}, etc.
Our paper also applied those three models to generate the adversarial samples under our framework.
FGSM generates the adversarial sample with one step perturbation along the gradient direction, and the step size is constrained to $\varepsilon$~\cite{goodfellow2014explaining}. The process is described as:
\begin{equation}
x_{adv} = x+\varepsilon \cdot \sign (\nabla_{x}l(h(x),y))
\end{equation}
PGD subdivides the process into several small steps. And after each step of perturbation, all the adversarial samples are projected to the $\varepsilon$-ball again~\cite{madry2018towards}. The process is described as:
\begin{equation}
x^{t}_{adv} = \Pi_{\varepsilon} (x^{t-1}+\beta\cdot \sign (\nabla_{x}l(h(x^{t-1}),y)))
\end{equation}where $\beta$ is the step size, and $\Pi(\cdot)$ represents the projection function.
Generally, PGD is the strongest first-order attack.
One of another popular attack method is C\&W.  The key idea of this attack method is to minimize the distance between normal samples and adversarial samples and to maximize the probability of the misclassification at the same time~\cite{carlini2017towards}. And the attack process is described as follows:
\begin{equation}
\begin{aligned}
d_{n} = \frac{1}{2}(\text{tanh}(w_{n})+1)-x_{n}\\
\underset{w_{n}}{\min}\|d_{n}\| + c\cdot f(\frac{1}{2}(\text{tanh}(w_{n})+1))\\
\end{aligned}
\end{equation}where $f(x')=\max(\max\{Z(x')_{i}:i\neq t\}-Z(x')_{t},-k)$ with $Z(x')_{t}$ represents the probability of the adversarial sample $x'$, $k$ represents the confidence, and $d_{n}$ is the distance between the normal sample and the adversarial sample.  
In this paper, to fully understand our proposed attack framework, we apply all three methods to the attack.

\section{Method Description}
\label{sec:methods}
\begin{figure}[htp]
  \centering
  \includegraphics[width=0.45\textwidth]{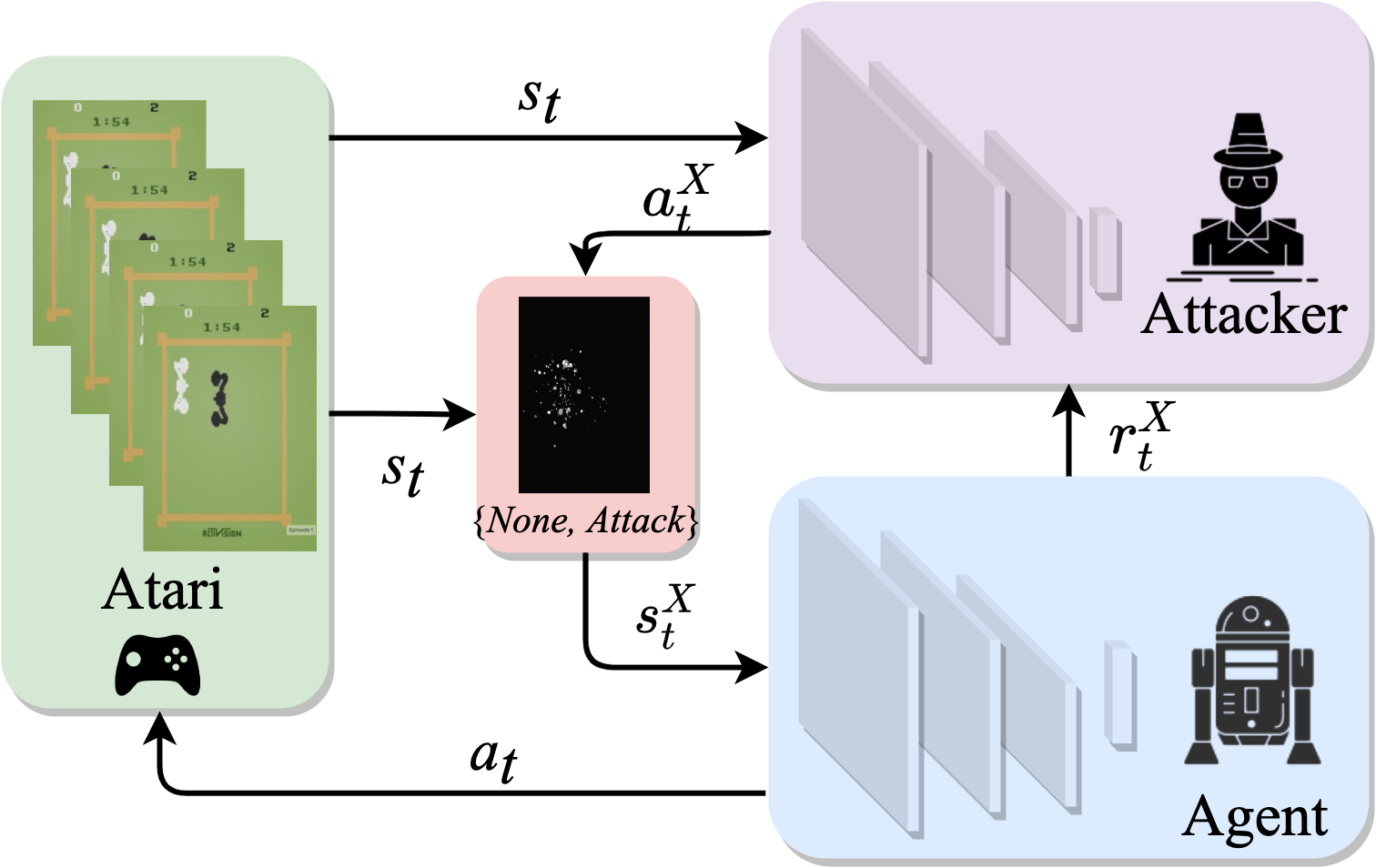}
  \caption{The structure of the proposed model.}
  \label{fig:structure}
\end{figure}

In this paper, we focus on the white box attack after DRL is deployed, and try to deploy the attack with an intelligent agent which is named attacker by learning a stable policy network $\theta^{X}$. 
With policy network $\theta^{X}$, the attacker will know when to attack by learning a certain attack pattern over observations.
The structure of the proposed model is shown in Figure~\ref{fig:structure}.
It is the man-in-the-middle format for the attacker when it deploys the attack on state $s_{t}$. Then the adversarial image will be sent to the agent directly, the reward $r^{X}$ of the attacker is originated from the agent. 
Based on this setting, the attacker's action space can be \{\textit{None, Attack}\}, noted that \textit{None} represents attacker not taking any action, and \textit{Attack} represents attacker adding perturbation on observation.
As we have mentioned above, most methods deploy the attack with the critical point to decide when conducting the attack~\cite{lin2017tactics,kos2017delving}, and this method is only based on rules and does not consider the long term reward, and is hard to obtain a good attacking performance. 
As we can see from the action space, the agent will be more stealthy if it takes more \textit{None} operations. On the contrary, it is not.
Generally, the goal of the model is to decrease DRL's performance with the fewest \textit{Attack} steps.
However, it is difficult to make the balance between attack effectiveness and stealthy. 
Based on previous work~\cite{sutton2018reinforcement}, the goal of DRL is to maximize the expected value of the cumulative sum of a received scalar reward signal.
Therefore, an effective way to guide the attacker is to design a combination of rewards, including effectiveness and stealth.
To cope with this problem, we have designed two types of rewards to help the attacker learn a better policy network $\theta^{X}$, and we will introduce them one by one.

\subsection{Short-Term Attacking Reward}
Intuitively, if an agent is attacked and the attack is successful, its action will immediately deviate from its original action, otherwise, we will consider the attack a failure. 
To encourage a successful short-term attack, if the new action taken from the disturbing observation is different from the original action, we will give a positive reward to the agent.
% it will be regarded as a successful attack.
An important feature of this reward is that it changes over time. Moreover, the ultimate goal is to ensure that all attacks have a successful outcome. Therefore, we can use the success rate of the attack to represent the short-term return.
In the long run, the success rate $r_{s}$ of the attack can be defined as: 
\begin{equation}
\label{equ:att}
r_{s} = \left \lfloor \frac{N_{a}}{N_{s}}  \right \rfloor
\end{equation}
where $ \left \lfloor \cdot \right \rfloor$ is the floor function to obtain one decimal place,
$N_{a}$ denotes the number of the successful attack before time $t$, and $N_{s}$ is the total number of the attack before time $t$.

As mentioned above, it is best to implement stealth bonus in a short-term reward. This can be indicated by the attacking ratio. If the ratio of \textit{None} is low, then the attacker has a low attack rate, and this can be considered an attack with good stealth. And the attack rate $r_{a}$ is defined as:
\begin{equation}
\label{equ:ra}
r_{a} =\left \lfloor  \frac{N_{s}}{N_{t}} \right \rfloor 
\end{equation}
where $N_{t}$ denotes the total play of the agent before time $t$.
From the work~\cite{sun2020stealthy}, we can conclude that an anomaly detector will be difficult to detect the attack when the attack number is small enough in deep reinforcement learning.
Therefore, to ensure its stealth, we need to ensure the attacking number is low enough.

To balance the attack effectiveness and stealth, the reward function for attacker at time $t$ is defined as:
\begin{equation}
\label{equ:reward}
r_{str} = r_{s} - r_{a}
\end{equation}
This reward is also named a short-term reward because it shows the immediate effect of the agent performing the action.
Different from previous work, we define the attack as a learning process where attackers need to learn the place (i.e., step) and the way (i.e., Attack on the action and attack on the state) to attack.
Furthermore, suppose the attack success rate is $100\%$ (i.e., $N_{a}=N_{s}$), and the agent play a oracle game which has $N_{s}=R_{a}$, then short-term reward will be $\frac{R_{a}-N_{s}}{R_{a}}$. From this function, we can conclude that the short-term reward is to ensure a minimal number of attacks by the attacker.

\subsection{Long-Term Attacking Reward}
As we have described, there is no connection between the deviated action and its original action in the short-term attack, and the attacking action only depends on the current state, that is to say, there is no connection between the current attack and the future reward.
However, to make up this problem, we need to find a good way to add the long-term effects to the attacker. One direct way is to apply the final rewards decrease to represent, and ensure the attack can accomplish some mission destruction tasks.
Therefore, the long-term rewards is defined as followed: 
\begin{equation}
    \label{equ:long_term_reward}
    r_{ltr} = \left \lfloor  \frac{R_{u}-r}{R_{u} - R_{l}} \right \rfloor
\end{equation}
If there are long-term effects over the agent, its final rewards will have a large decrease which can be represented by the percentage decline of the rewards. $r$ denotes the accumulate reward of the current step $r = \sum_{i}^{t}r_{i}$, where $r_{i}$ is the normal reward in the original setting.
$R_{u}$ is the upper bound of the accumulate reward, and $R_{l}$ denotes the lower bound the accumulate reward.
In our framework, the final reward is defined as a summation format with different ratio, i.e.,
\begin{equation}
\label{equ:r_total}
r_{t}^{X} = \alpha r_{str} + (1 - \alpha) r_{ltr}
\end{equation}
Where $\alpha$ is the hyperparameter that defines the ratio of $r_{str}$ and $r_{ltr}$.

In general, the short-term reward is designed to stimulate agents to achieve attacks with fewer attacks which in turn ensures the stealthiness of the attacks, while the long-term reward is designed to ensure the effectiveness of the attacks by minimizing the accumulated rewards. 

And the way to train the attacker $X$ is designed as shown in Algorithm~\ref{alg:1} by freezing the well-trained agent $T$.
In the algorithm, $M$ is the number of epochs, $D$ is the replay memory of the attacker, and its capacity is set to $N$.

\begin{algorithm}[htp] 
\small
 \caption{The attacking against the reinforcement learning with man-in-the-middle.}
 \label{alg:1}
%  \begin{algorithmic}[1]
\KwIn{Environment Atari, Attacker $X$, Agent $T$; \\Initialize replay memory $D$ to capacity $N$;\\ Initialize $X$'s parameters $Q(s^{X},a^{X})$ with random weight $\theta^{X}$ and load the well-trained $T$'s weight $\theta^{T}$;\\
Initialization target action-value function $\hat{Q} $ of attacker with  $\theta^{X}_{-}=\theta^{X}$;}
 \For {episode=1,M}
 {Initialize status $s_{t}$; \\
  \While{not \textit{Done}}
    {
    Select Agent $T$'s action $a_{t}$;\\
    Sample a value $v$ from a uniform distribution $[0,1]$;\\
    \eIf{$v\leq \epsilon$}
      {
      Select a random action $a^{X}_{t}$ from attack action space;\\
      }
      {select $a_{t}^{X}=\argmax_{a^{X}}Q(s^{X}_{t},a_{t}^{X};\theta^{X})$;\\
      } 
%       \STATE Execute action $a^{X}_{t}$ and observe reward $\hat{r}_{t}$, and status $s_{t+1}$
%       \STATE Store transition ($s_{t},a_{t},r_{t},s_{t+1}$) in D
      \eIf{$a_{t}^{X}= \textit{None}$}
        {
        Set $s_{t}^{X} = s_{t} $;
        }
       {
       Obtained $s_{t}^{X}$ with attack methods $ATK$;\\
       Select new action $a_{t_{new}} = \arg\max_{a^{X}}Q(s_{t}^{X},a_{t};\theta^{T})$;\\
           \If{$a_{t_{new}}!=a_{t}$}
           {
           Attack Success;
           }
       }
       Obtain $r^{X}_{t}$ with Equation~\ref{equ:r_total};\\
    %   \IF{Done}
    %   \STATE Adding long-term attack rewards $r^{X}_{l}$ to $r^{X}_{t}$
    %   \ENDIF
      Store transition ($s^{X}_{t},a^{X}_{t},r^{X}_{t},s^{X}_{t+1}$) in D;\\
      Sample random minibatch of transition ($s_{j},a_{j},r_{j},s_{j+1}$) from $D$;\\
      Set $y_{t}=\left\{\begin{matrix}  r_{j}&\text{\makecell{terminates at $j+1$}}\\ r_{j}+\gamma\max_{a'}\hat{Q}(s_{j+1},a';\theta^{X}_{-})  & \text{otherwise}  \end{matrix}\right.$;\\
      Perform a gradient descent step on $(y_{j}-Q(s_{j},a_{j};\theta^{X}))^{2}$ w.r.t. the parameter $\theta^{X}$;\\
     Set $s_{t+1}=s_{t}$;\\
     Every $C$ steps reset $\hat{Q}=Q$;\\
    }
 }
 
\end{algorithm}
In the algorithm, the whole process is represented by lines 2-26, where in line 2 we need to initialize the agent state $s_{t}$ at the beginning of each episode.
Line 4 is the action selection of the trained agent $T$. Lines 5-10 indicate action selection by the attacker $X$ using the $epsilon$-greedy strategy.
Lines 11-19 represent the attacking steps in the training, where lines 11-12 represent that the attacker will do nothing when the action is $None$, lines 13-19 represent the procedures when action is not $None$, where line 14 is the attack with the $ATK$ method, and when the attack is completed, a new state will be obtained and on which the agent will have a new action, which is in line 15.
And lines 16-18 are to determine whether the attack is successful.
Line 20 is to obtain the reward after the attack with Equation~\ref{equ:r_total}, and line 21 is to store the transition sample into memory $D$ for rollout training. Lines 22-24 indicate the model optimization procedures. Line 25 indicate updating the target action-value function $\hat{Q}$ with $Q$ every $C$ steps.

\section{Experiments}
 \label{sec:experiments} 
 In the experiment, we will first verify the effectiveness of the proposed framework in the attack, then show the transferability of the attacking framework in different scenarios, i.e., transfer across the environment and transfer across the attacking methods, and finally apply adversarial training to enhance the robustness of DRL.
 
\subsection{Settings}
\begin{figure}[htp]
    \centering
    \includegraphics[width=0.5\textwidth]{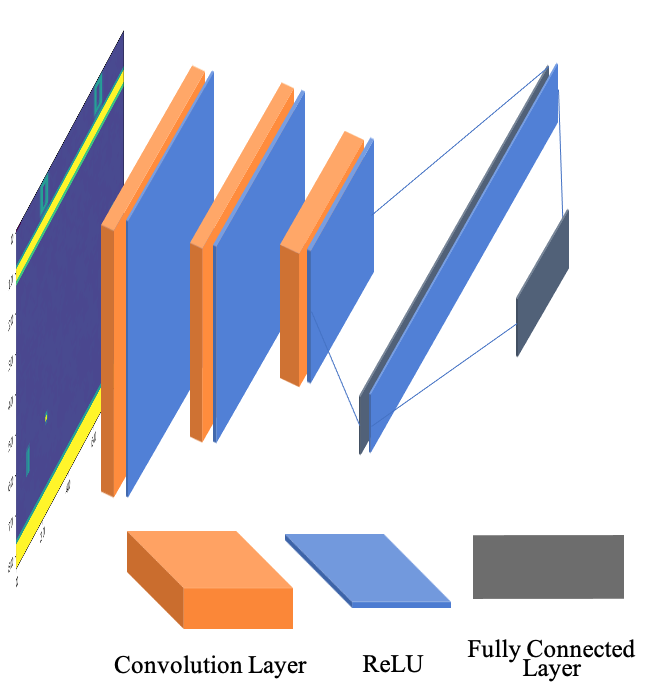}
    \caption{The structure of the policy network}
    \label{fig:policynetwork}
\end{figure}

All of our experiments are run on the gym platform. The games selected include $\{$\textit{Pong}, \textit{Fishing Derby} (\textit{Fishing} for short), \textit{Boxing}, \textit{Seaquest}, \textit{MsPacMan} (\textit{PacMan} for short)$\}$. 
The DRL is implemented in python based on Pytorch package~\footnote{https://pytorch.org/}, and the input of Agent and the attacker from Atari's games are selected image frame which is transferred to 84$\times$84-pixel image, and the policy network is a classical convolutional neural network which is to map the input to the action space. There are three convolutional layers with the size of (32, 8, 8, 4), (64, 4, 4, 2) and (64,  3, 3, 1), where the first value in the bracket is the number of the filters, the second and the third values denote the filter size, and the last value represents the stride size. The last two layers are the fully connected functions that map the hidden representation to the action space, and the shape of the weights in these two layers are (3136, 512) and (512, action space). As we have described earlier, there are two actions for the attacker which are $\{None, Attack\}$. The activation function is ReLU. And there is the same structure of the policy neural network in Attacker.
The structure of the policy network is illustrated in Figure~\ref{fig:policynetwork}.

The detailed hyper-parameters are listed in Table~\ref{tab:paras}.
Specifically, the learning rate is 1e-4, and the batch size is 32. The replay memory buffer size is set to 1e6, and the discount factor $\gamma$ is 0.99. During the training, the current network updating frequency is 4, and the target network updating frequency is 100 by copying the parameters of $Q(s^{X},a^{X})$ to the target network (i.e., $C=100$). In the $\epsilon$-greedy policy, the start value is $1.0$, and the end value is 0.01 with the fraction of number steps 0.1. The training starts after 1e4 steps, and the maximum training step is 1e7. To balance $r_{str}$ and $r_{ltr}$, $\alpha$ is set to 0.5 in Equation~\ref{equ:r_total}.

\begin{table*}[htp]
  \centering
\begin{tabular}{l|l|l|l}
\hline
Parameter      & Value & Parameter  & Value \\\hline
learning rate &    1e-4   & batch size &  32     \\\hline
replay buffer $D$ &    1e6    & update current network frequency &  4     \\\hline
discount factor $\gamma$    &   0.99    &   update target network frequency $C$ &  100    \\\hline
$\epsilon$-greedy start threshold    &   1.0    & train stage  &  >1e4    \\\hline
$\epsilon $-greedy end threshold    &  0.01    & fraction of number steps  &  0.1    \\\hline
$\alpha$   &  0.5    &   maximum number steps     &  1e7    \\\hline
% $\alpha$   &  0.5    &        &     \\\hline
\end{tabular}
\caption{The hyper-parameters applied in this paper.}
\label{tab:paras}
\end{table*}

The evaluation metrics include the designed long-term reward  $r_{ltr}$ and short-term reward $r_{str}$.
To make fair comparisons, all the experiments are sharing the same settings, and all the results are averaged from three times test.
We will report the performance of different attacking methods, and the detailed models are listed as followed:
\begin{itemize}
\item \textbf{Uniform}
Uniform denotes attacks are deployed in a uniform distribution. And this was proposed by Huang et al.~\cite{huang2017adversarial}.
\item \textbf{Strategically Timed} Strategical Timed means the attack only happens at the point where the attack score is higher than the threshold, attack score can be calculated as in Equation~\ref{equ:thres1}. And this was proposed by Lin et al.~\cite{lin2017tactics}.
\end{itemize}
 
\subsection{Attack Evaluation}
To evaluate the attacking effectiveness, we first train an oracle agent $T$ on the games we selected. Then train the attacker $X$ by freezing $T$. 
In order to have a good understanding of the attacking stealthy and effectiveness, all the results are reported in the designed balanced rewards $r^{X}$ that are calculated in Equation~\ref{equ:r_total}. And the results are shown in Table~\ref{Tab:sres}.

\begin{table*}[htp]
\centering
\begin{tabular}{c|c|c|c|c|c}
\hline
Methods & Pong & Seaquest & Boxing  & Pacman & Fishing \\\hline
Uniform(FGSM) &  0.77  &  0.67 &  0.66   & 0.64   &   0.70  \\ 
Uniform(PGD) & 0.81  &   0.64     &   0.70      &   0.67       & 0.70  \\ 
Uniform(C\&W) & 0.04   &    0.54    &    0.37    &  0.56      &  0.54  \\\hline
Strategically Timed(FGSM) &  0.86 & 0.74   & 0.79    &  0.59   & 0.62 \\
Strategically Timed(PGD) &  0.87 & 0.74   & 0.89    &  0.65   & 0.66 \\
Strategically Timed(C\&W) &  0.07 & 0.62   & 0.46    &  0.40   & 0.47 \\\hline
% Critical Point & 0.32   &      &     &     &   \\\hline
Our(FGSM)  & \textbf{ 0.85}    & \textbf{0.80}  &   \textbf{0.86}   &    \textbf{0.78}   &  \textbf{0.72}  \\
Our(PGD)  &  \textbf{0.98}    &\textbf{ 0.81}  & \textbf{  0.92 }  &    \textbf{0.82}  &  \textbf{0.75}  \\
Our(C\&W)  &  \textbf{0.15}    & \textbf{0.69}  &   \textbf{0.51 } &    \textbf{0.56 }  &  \textbf{0.54}  \\\hline
\end{tabular}
\caption{The attack evaluation with different attacking models, the higher the better, the value is calculated with Equation~\ref{equ:r_total}.}
\label{Tab:sres}
\end{table*}

\begin{figure}[htp]
\centering
\includegraphics[width=0.49\textwidth]{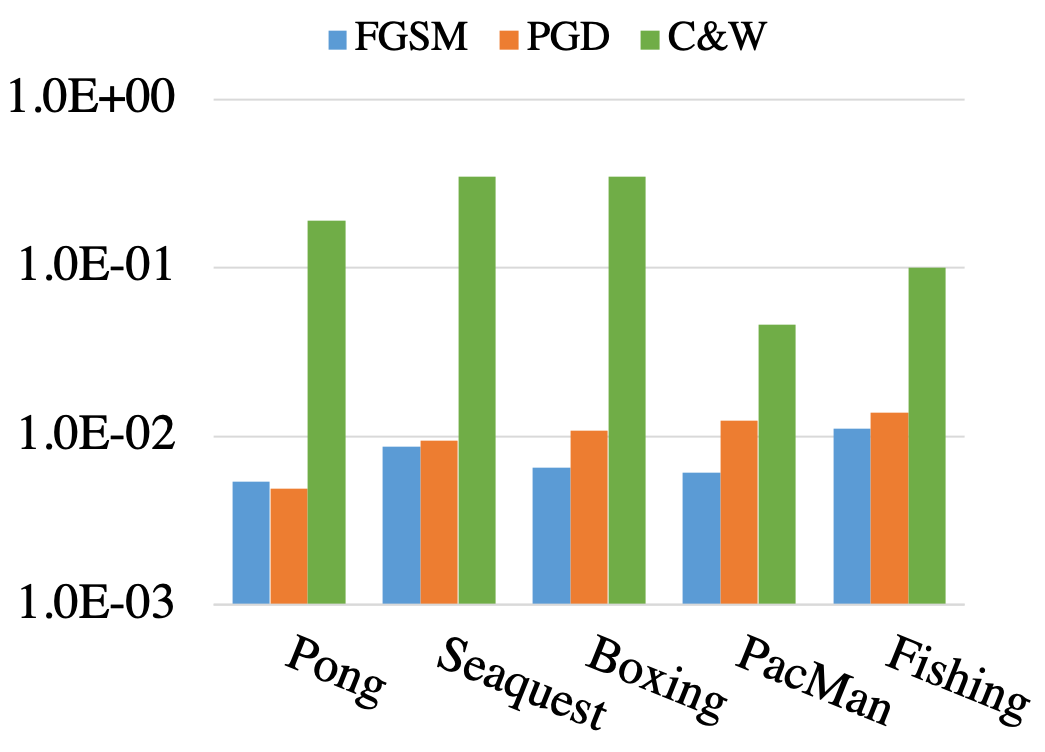}
\caption{The Time with different attack approaches.}
\label{fig:time}
\end{figure}

\begin{figure*}[htp]
\centering
\subcaptionbox{FGSM (1.78)\label{subfig:lef}}
{\includegraphics[width=0.32\textwidth]{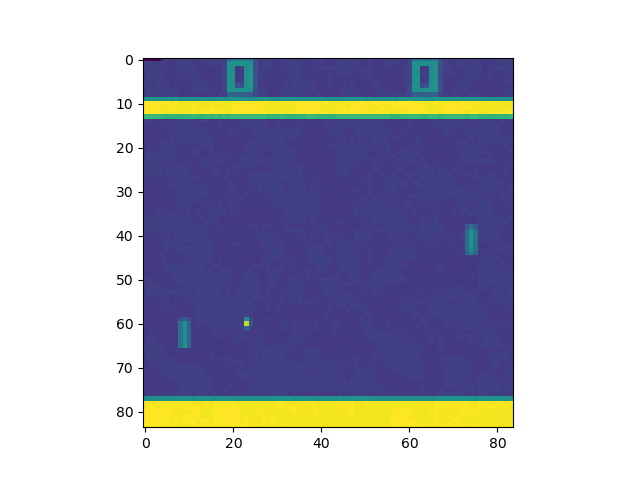}}
\subcaptionbox{PGD (1.71)\label{subfig:mid}}
{\includegraphics[width=0.32\textwidth]{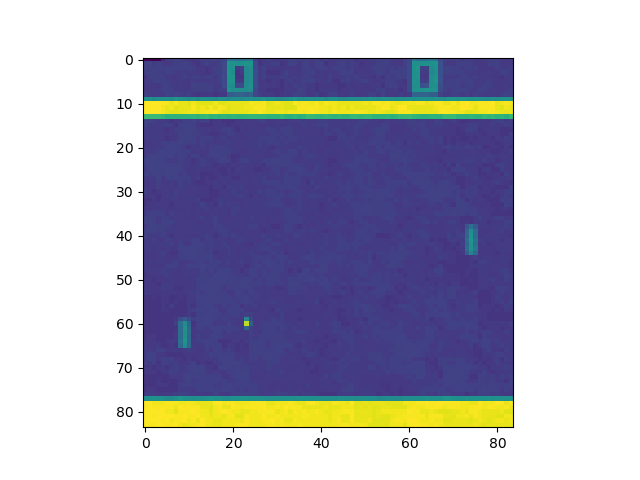}}
\subcaptionbox{C\&W (0.66)\label{subfig:rig}}
{\includegraphics[width=0.32\textwidth]{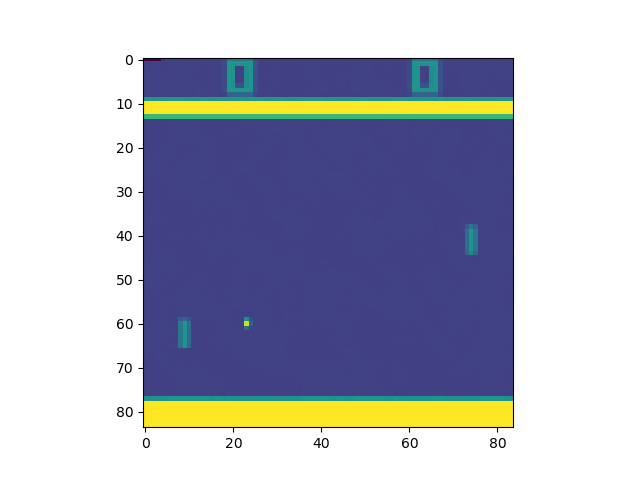}}
\caption{The craft adversarial frame with different attack approaches, the value in the bracket indicates the absolute distance from the original frame.}
\label{fig:advimage}
\end{figure*}

From the table, we can see that all the cases in our proposed framework have a better performance as it has higher attacking rewards. To our surprisingly, C\&W is not the best-attacking model, on the contrary, and its performance is the worst one compared with the other two attacking approaches.
One reason we think is that the attacking agent can not learn a stable policy when facing the small state variance, as one of the objectives of C\&W is to minimize the distance between the adversarial sample and the original sample as such it is a model-specific attack algorithm, therefore its use here makes its attack performance limited. 
To validate this hypothesis, we compared images that were attacked by different models at the same time, and these results are shown in Figure~\ref{fig:advimage}. In the meantime, we also reported the time usage with different attacking approaches, and the results are shown in Figure~\ref{fig:time}.
From Figure~\ref{fig:advimage}, we can see that the adversarial sample from C\&W is the cleanest, and from Figure~\ref{fig:time}, C\&W has the longest time for attack conducting.
And these make it difficult for the attacker to learn the adversarial strategy because of the long-time adversarial sample generation and the little change of the adversarial sample generated.
However, under our proposed framework, PGD achieves the best performance. And this is also validated in the cases of ``Strategically Timed'' and ``Uniform''.

\subsubsection{Short-term Reward \& Long-term Reward}
\begin{figure*}[htp]
\centering
\includegraphics[width=0.89\textwidth]{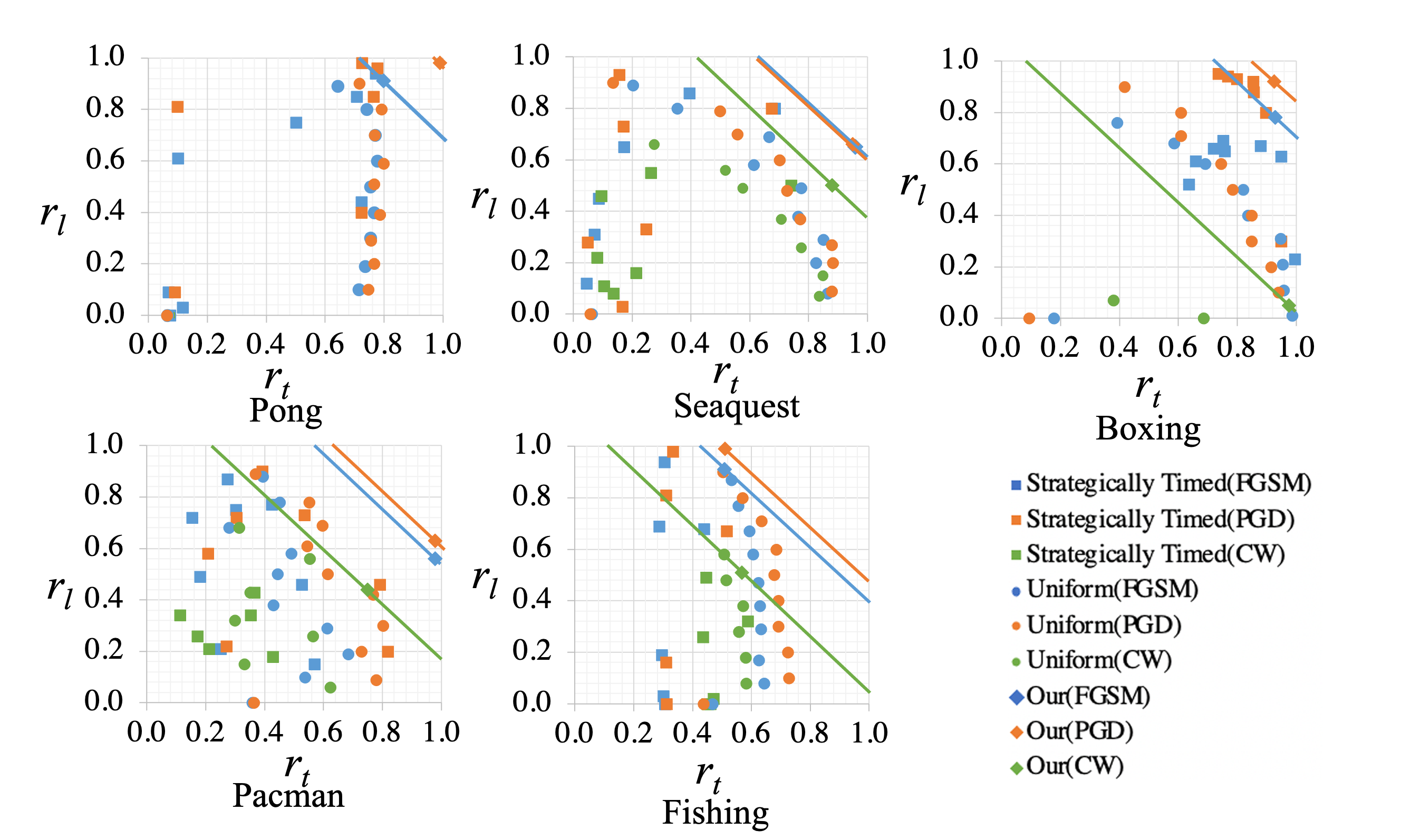}
\caption{The detailed results.}
\label{fig:res2}
\end{figure*}

To make the results more clear, we use both short-term reward and long-term reward to demonstrate the attack reward $r^{X}$. 
The short-term attack denotes the single attack affects the agent's behavior directly, i.e., high attacking success rate with the low attacking rate. And the long-term attack is the embodiment of the final effect of the series of attacks. 
And the results are shown in Figure~\ref{fig:res2}, and the points on the top right will have better performance in both immediate and long-term attacks.
% Therefore, we use the attacking rewards as the evaluation metric in this section, and the results are shown in Table~\ref{Tab:short_term}.

In the figure, different attack strategies are shown in different shapes. $\sbullet[2]$ denotes the ``Uniform'' attack, $\blacksquare$ is the ``Strategically Timed'' attack, and $ \rotatebox[origin=c]{45}{$\blacksquare$}$ is our proposed attacking framework. 
And points of the same color represent the same type of attack: blue is the FGSM, brown is the PGD and green is the C\&W.
The straight lines are the contours that have the same $r^{X}$. 
From the figure, we can see that our proposed framework has the best performance as all nodes of the same type are below these lines. 
And especially in the game of Pong, the attack with PGD under our proposed framework has the best performance by achieving the highest score in long-term reward and short-term reward at the same time. 
And this means our proposed framework has obtained the ability of effective stealthy attack.
Specifically, there is a negative correlation between short-term reward and long-term reward, and the relationship between the two is difficult to balance, and this phenomenon is validated both in the cases of ``Strategically Timed'' and ``Uniform''.
Another interesting discovery in these two attacking scenarios is that all the games prefer having higher long-term rewards no matter what short-term rewards are, especially when the long-term reward is higher than the threshold of 0.5. 

\subsubsection{Advantages of the Metrics}
 In this paper, we applied the newly designed metrics to evaluate the attack effectiveness and stealthy, to validate the wealth and goodness of the proposed new metrics, a comparison is made over the Pong game whose upper bound of the reward is 21, and the results are illustrated in Table~\ref{tab:metric}.
 \begin{table}[htp]
 \centering
\begin{tabular}{c|c|c|c|c|c|c}
\hline
Case & Rewards  &  $r_{s}$ & $r_{a}$&  $r_{str}$  & $r_{ltr}$    & $r$ \\\hline
1 &-9.38     & 100\%    & 56\%   & 0.44 & 0.72 & 0.58 \\\hline
2 &-8.69     & 100\%    & 15\%    & 0.85 & 0.71 & 0.78 \\\hline
3 &-11.44    & 98\%  &  4\%    & 0.94 & 0.77 & 0.86\\\hline
\end{tabular}
\caption{The comparisons between different evaluation metrics.}
\label{tab:metric}
\end{table}
The general way to show the effectiveness of an attack is to use the reduction of the reward, which can be seen in the first column of Table~\ref{tab:metric}. This approach is straightforward and clear in terms of attack effectiveness measurement and is used in many works ~\cite{sun2020stealthy,zhao2020blackbox}. However, we first need to know the upper bound of the reward and then can know how much the attack is reduced, meanwhile, this measurement does not reveal how many attacks have been launched against the agent, for example, in cases 1 and 2, the attack success ratio $r_{s}$ against the agent is 100\% in both cases, but the attack ratio $r_{a}$ differs significantly. 
However, among our evaluation metrics, there are attack success ratio $r_{s}$ and attack ratio $r_{a}$, which directly yield the short-term reward $r_{str}$, and this is a good metric to evaluate stealthiness. Therefore, in terms of our last column of Table~\ref{tab:metric}, our proposed metric $r$ balances effectiveness and stealthy well.

\subsection{Transferabilities}
How does the well-trained attacker behave when transferred to another environment,
and how does the well-trained policy performs when we cooperate with another $ATK$ (i.e., FGSM, PGD, and C\&W).
To answer these two questions well, we first transfer the well-trained attacker to different environments (i.e., games), and the details are presented in Section~\ref{sec:envtrans}. We then cooperate the well-trained policy with different attacking methods, and the details are described in Section~\ref{sec:transfer_attack}.

\subsubsection{Transferability Across Environments}
\label{sec:envtrans}
In this scenario, we select PGD which performs the best as the attacking method. Then we transfer the well-trained policy to other environments, and the rewards after attacking are reported in Table~\ref{tab:trans_env}. 

\begin{table}[htp]
\centering
\begin{tabular}{cc|c|c|c|c|c}
\cline{3-7} 
 & \multicolumn{1}{l}{}         & \multicolumn{5}{c}{From}                        \\ \cline{3-7} 
\multicolumn{1}{l}{}  &  \multicolumn{1}{l}{}  & Pong & Sequest & Boxing & PacMan & Fishing \\\hline
\multicolumn{1}{c|}{\multirow{5}{*}{To}}  & Pong    & 0.98 & \cellcolor{red!95} 0.04 & \cellcolor{red!62} 0.37 &\cellcolor{red!45} 0.54 &\cellcolor{red!95} 0.04    \\\cline{2-7} 
\multicolumn{1}{c|}{} & Sequest &\cellcolor{red!41} 0.48 & 0.81    & \cellcolor{red!37}0.51   & \cellcolor{red!35}0.52   & \cellcolor{red!41}0.48    \\\cline{2-7} 
\multicolumn{1}{c|}{} & Boxing  &\cellcolor{red!49} 0.47 &\cellcolor{red!89} 0.10  & 0.92   & \cellcolor{red!35}0.60   &\cellcolor{red!89} 0.10     \\\cline{2-7} 
\multicolumn{1}{c|}{} & Pacman  &\cellcolor{red!38} 0.51 &\cellcolor{red!41} 0.48  & \cellcolor{red!39}0.50   & 0.82   & \cellcolor{red!41}0.48    \\\cline{2-7} 
\multicolumn{1}{c|}{} & Fishing & \cellcolor{red!15}0.64 & \cellcolor{red!80}0.15    & \cellcolor{red!61}0.29   &\cellcolor{red!53} 0.35   & 0.75   \\\hline
\end{tabular}
\caption{Rewards $r^{X}$ based on different transfer cases, ``From'' represents the original environment in which the attack model was trained, and ``To'' represents the game to which the attack model was transferred. The background color indicates the degree of the decrease compared to the normal condition. The darker the color, the greater the degree.}
\label{tab:trans_env}
\end{table}

From the table, we can see that almost all the attackers have a performance decrease compared with their original environment, and the performance of some cases drops as much as 90\%.  
The worst case is the policy learned from Seaquest and Fishing, and they can hardly adapt in the new environment as more than half of the cases are in low reward $r^{X}$.
However, there is still some transferring capability in our model. For example, with strategies learned from Pong, Boxing, and PacMan, the performance across different environments remains around 0.5. 
Based on those observations, we can see that when the attacker enters a new environment, it still needs to be fine-tuned or retrained to adapt to the new state. 

\subsubsection{Transferability Across Attacking Methods}
\label{sec:transfer_attack}
Generally, there are three well-trained policies under three attacking methods. To validate the transferability of the well-trained policy across different attacking methods and the attacking methods across different well-trained policies, we select the Seaquest as the test-bed, and we change different attacking methods described in line 14 in Algorithm~\ref{alg:1}.
And the experiment results are reported in Table~\ref{tab:trans_policy}.

\begin{table}[htp]
\centering
\begin{tabular}{ccccc}
\cline{3-5}
\multicolumn{1}{l}{}                         & \multicolumn{1}{l}{}      & \multicolumn{3}{c}{$ATK$}                        \\ \cline{3-5} 
\multicolumn{1}{l}{}                         &                           & \multicolumn{1}{c|}{FGSM} & \multicolumn{1}{c|}{PGD}  & C\&W \\ \hline
\multicolumn{1}{c|}{\multirow{3}{*}{Policy}} & \multicolumn{1}{c|}{FGSM} & \multicolumn{1}{c|}{0.80}  & \multicolumn{1}{c|}{\cellcolor{red!5}0.76} & \cellcolor{red!20} 0.64 \\ \cline{2-5} 
\multicolumn{1}{c|}{}                        & \multicolumn{1}{c|}{PGD}  & \multicolumn{1}{c|}{\cellcolor{green!0} 0.81} & \multicolumn{1}{c|}{0.81} & \cellcolor{red!11}0.72 \\ \cline{2-5} 
\multicolumn{1}{c|}{}                        & \multicolumn{1}{c|}{C\&W} & \multicolumn{1}{c|}{\cellcolor{green!7}0.75} & \multicolumn{1}{c|}{\cellcolor{green!3}0.71} & 0.69 \\ \hline
\end{tabular}
\caption{Rewards $r^{X}$ based on different transfer cases, ``Attacking Methods'' are selected from {FGSM, PGD, C\&W}, and ``Policy'' is the well-trained policy with different attacking methods.  background color indicates the degree of the increase and decrease compared to normal condition, with green indicating the increase rate and red indicating the decrease rate. The darker the color, the greater the degree.}
\label{tab:trans_policy}
\end{table}
In the table~\ref{tab:trans_policy}, with the FGSM learning the policy, the performance degradation degree increases successively when the attacking method is changed to PGD and C\&W, and when using the PGD learning the policy, changing the attacking method to C\&W will also degrade the performance, while changing the attacking method to FGSM does not. However, when the attacking method is changed to PGD and FGSM, the performance is improved successively by using the C\&W learning strategy.
From these observations, it can be seen that the transferability of the attacking policy across different attacking methods is not as sensitive as the environment. 
The same applies to the transferability of the attacking methods across the different attacking policies.
Compared with the simple attacking method (e.g., FGSM), the policy learned from a more complex attacking method (e.g., C\&W) has better transferability. 
And the attack method PGD also has competitive performance compared with C\&W, which is one of the reasons why we can get the best performance by using PGD in our framework.

\subsection{Adversarial Training}
To validate the robustness of the model, we obtain the robust model (i.e., $T_{R}$) by fine-tuning the oracle DQN $T$ with the well-trained attacker $X$ that attacks with PGD, then validate the robustness with different attacking methods. The procedures are shown in Algorithm~\ref{alg:2}. And the experiment results are listed in Table~\ref{Tab:res3}.

\begin{algorithm}[H] 
 \caption{The adversarial training procedure.}
 \label{alg:2}
%  \begin{algorithmic}[1]
    \KwIn{Environment Atari}
    Train an oracle agent $T$.\\
    Freeze agent $T$, training attacker $X$ with PGD.\\
    Freeze the well-trained attacker $X$, fine-tune oracle agent $T$ with adversarial samples to obtain the robustness model $T_{R}$.\\
\end{algorithm}

\begin{table}[htp]
\centering
\begin{tabular}{c|c|c|c|c|c}
\hline
    & Pong    & Fishing & Boxing  & Seaquest & PacMan \\\hline
$T$   & 17.91   & 30.63   & 69.56   & 1177.50   & 804.69   \\\hline
$T^{R}$           &\cellcolor{green!10} 19.87    & \cellcolor{green!2} 31.34  & \cellcolor{green!7}74.66 & \cellcolor{green!19} 1400.00     &\cellcolor{green!20}  1020.00    \\\hline
Attack (FGSM) &\cellcolor{green!4} 18.59 & \cellcolor{red!11}   27.22 &\cellcolor{red!42} 40.65 &  \cellcolor{red!20} 980.00       &  \cellcolor{red!27}   587.19     \\\hline
Attack (PGD)  & \cellcolor{green!11}19.94 & \cellcolor{red!1} 30.06    & \cellcolor{red!16}58.56 & \cellcolor{red!8}1080.00     & \cellcolor{red!18}660.00      \\\hline
Attack (C\&W)   &\cellcolor{green!7} 19.24 & \cellcolor{red!66} 10.31  &\cellcolor{red!48} 36.34 &\cellcolor{red!19} 950.00      &\cellcolor{red!42} 464.06\\\hline
\end{tabular}
\caption{Adversarial training with the well-trained adversarial attack. The background color indicates the degree of the increase and decrease compared to normal condition, with green indicating the increase rate and red indicating the decrease rate. The darker the color, the greater the degree.}
\label{Tab:res3}
\end{table}

From the table, we can see that our adversarial training has boosted the model performance, and the improvement is ranging from 2.32\% to  18.90\%. That means adversarial training with the adversarial samples compensates for the vulnerability of the model to some degree. 
When we attack the robust agent $T_{R}$ again, we can see that there is still a performance decrease in most of the cases, but the degradation is less than the attack that deployed directly on the Oracle agent $T$.
In particular, the performance degradation of the PGD attack is the lowest, followed by the FGSM attack, and the performance degradation of the C\&W attack is the largest.
From these observations, we can see that the robustness of the model has indeed been improved to a certain extent after adversarial training, and the reason why PGD attack has the lowest performance degradation is that we used it to train the adversarial policy.

\section{Conclusions \& Future Works}
\label{sec:conclusions}
In this paper, we proposed an intelligent attacker with reinforcement learning by considering the short-term attack, and long-term attack at the same time, then provided a useful evaluation metric to measure the attacking effectiveness and stealth.
In our paper, the attack was deployed in a white-box environment where we know the actions of the target agent, as well as information about the state of each frame after the agent, performs the action, and this prior information can be easily obtained since the state and action are explicit information of the target agent.
From the experimental results, we can see that the proposed attack method can achieve the goal in short steps attack. In the following step, we also conducted adversarial training to improve the robustness of reinforcement learning.  
In future work, we will explore more intelligent attack agents and try to explain why such attacks are useful in black-box scenarios, where one can employ parallel agents as target agents to accomplish policy network learning for the attack. Furthermore, we will explore how the robustness of the model under adversarial training and uncover the impact on the overestimate which is the inherent shortcoming of the DQN when deploying attacks.

\section*{Acknowledgments}
This research is supported by the National Natural Science Foundation of China (No.62103330), and the Fundamental Research Funds for the Central Universities (3102021ZDHQD09). 
\bibliographystyle{plain}
\bibliography{references}

\begin{thebibliography}{10}

\bibitem{behzadan2019analysis}
Vahid Behzadan and William Hsu.
\newblock Analysis and improvement of adversarial training in dqn agents with
  adversarially-guided exploration (age).
\newblock {\em arXiv preprint arXiv:1906.01119}, 2019.

\bibitem{behzadan2017vulnerability}
Vahid Behzadan and Arslan Munir.
\newblock Vulnerability of deep reinforcement learning to policy induction
  attacks.
\newblock In {\em International Conference on Machine Learning and Data Mining
  in Pattern Recognition}, pages 262--275. Springer, 2017.

\bibitem{behzadan2017whatever}
Vahid Behzadan and Arslan Munir.
\newblock Whatever does not kill deep reinforcement learning, makes it
  stronger.
\newblock {\em arXiv preprint arXiv:1712.09344}, 2017.

\bibitem{behzadan2018mitigation}
Vahid Behzadan and Arslan Munir.
\newblock Mitigation of policy manipulation attacks on deep q-networks with
  parameter-space noise.
\newblock In {\em International Conference on Computer Safety, Reliability, and
  Security}, pages 406--417. Springer, 2018.

\bibitem{carlini2017magnet}
Nicholas Carlini and David Wagner.
\newblock Magnet and" efficient defenses against adversarial attacks" are not
  robust to adversarial examples.
\newblock {\em arXiv preprint arXiv:1711.08478}, 2017.

\bibitem{carlini2017towards}
Nicholas Carlini and David Wagner.
\newblock Towards evaluating the robustness of neural networks.
\newblock In {\em 2017 IEEE Symposium on Security and Privacy}, pages 39--57.
  IEEE, 2017.

\bibitem{chen2021survey}
JinYin Chen, Yan Zhang, XueKe Wang, HongBin Cai, Jue Wang, and ShouLing Ji.
\newblock A survey of attack, defense and related security analysis for deep
  reinforcement learning.
\newblock {\em Acta Automatica Sinica,}, 45:1--19, 2020.

\bibitem{chen2019adversarial}
Tong Chen, Jiqiang Liu, Yingxiao Xiang, Wenjia Niu, Endong Tong, and Zhen Han.
\newblock Adversarial attack and defense in reinforcement learning-from ai
  security view.
\newblock {\em Cybersecurity}, 2(1):1--22, 2019.

\bibitem{goodfellow2014explaining}
Ian~J Goodfellow, Jonathon Shlens, and Christian Szegedy.
\newblock Explaining and harnessing adversarial examples.
\newblock {\em arXiv preprint arXiv:1412.6572}, 2014.

\bibitem{gu2019adversary}
Zhaoyuan Gu, Zhenzhong Jia, and Howie Choset.
\newblock Adversary a3c for robust reinforcement learning.
\newblock {\em arXiv preprint arXiv:1912.00330}, 2019.

\bibitem{han2018reinforcement}
Yi~Han, Benjamin~IP Rubinstein, Tamas Abraham, Tansu Alpcan, Olivier De~Vel,
  Sarah Erfani, David Hubczenko, Christopher Leckie, and Paul Montague.
\newblock Reinforcement learning for autonomous defence in software-defined
  networking.
\newblock In {\em International Conference on Decision and Game Theory for
  Security}, pages 145--165. Springer, 2018.

\bibitem{huang2017adversarial}
Sandy Huang, Nicolas Papernot, Ian Goodfellow, Yan Duan, and Pieter Abbeel.
\newblock Adversarial attacks on neural network policies.
\newblock {\em arXiv preprint arXiv:1702.02284}, 2017.

\bibitem{kiourti2019trojdrl}
Panagiota Kiourti, Kacper Wardega, Susmit Jha, and Wenchao Li.
\newblock Trojdrl: Trojan attacks on deep reinforcement learning agents.
\newblock {\em arXiv preprint arXiv:1903.06638}, 2019.

\bibitem{kos2017delving}
Jernej Kos and Dawn Song.
\newblock Delving into adversarial attacks on deep policies.
\newblock {\em arXiv preprint arXiv:1705.06452}, 2017.

\bibitem{lee2020spatiotemporally}
Xian~Yeow Lee, Sambit Ghadai, Kai~Liang Tan, Chinmay Hegde, and Soumik Sarkar.
\newblock Spatiotemporally constrained action space attacks on deep
  reinforcement learning agents.
\newblock In {\em Proceedings of the AAAI Conference on Artificial
  Intelligence}, volume 34(04), pages 4577--4584, 2020.

\bibitem{li2019deep}
Haoran Li, Qichao Zhang, and Dongbin Zhao.
\newblock Deep reinforcement learning-based automatic exploration for
  navigation in unknown environment.
\newblock {\em IEEE transactions on neural networks and learning systems},
  31(6):2064--2076, 2019.

\bibitem{li2018generative}
Yang Li, Quan Pan, Suhang Wang, Tao Yang, and Erik Cambria.
\newblock A generative model for category text generation.
\newblock {\em Information Sciences}, 450:301--315, 2018.

\bibitem{li2021graph}
Yang Li, Wei Zhao, Erik Cambria, Suhang Wang, and Steffen Eger.
\newblock Graph routing between capsules.
\newblock {\em Neural Networks}, 143:345--354, 2021.

\bibitem{lin2017tactics}
Yen-Chen Lin, Zhang-Wei Hong, Yuan-Hong Liao, Meng-Li Shih, Ming-Yu Liu, and
  Min Sun.
\newblock Tactics of adversarial attack on deep reinforcement learning agents.
\newblock In {\em Proceedings of the 26th International Joint Conference on
  Artificial Intelligence}, pages 3756--3762, 2017.

\bibitem{madry2018towards}
Aleksander Madry, Aleksandar Makelov, Ludwig Schmidt, Dimitris Tsipras, and
  Adrian Vladu.
\newblock Towards deep learning models resistant to adversarial attacks.
\newblock In {\em International Conference on Learning Representations}, 2018.

\bibitem{neklyudov2018variance}
Kirill Neklyudov, Dmitry Molchanov, Arsenii Ashukha, and Dmitry Vetrov.
\newblock Variance networks: When expectation does not meet your expectations.
\newblock In {\em International Conference on Learning Representations}, 2018.

\bibitem{pattanaik2018robust}
Anay Pattanaik, Zhenyi Tang, Shuijing Liu, Gautham Bommannan, and Girish
  Chowdhary.
\newblock Robust deep reinforcement learning with adversarial attacks.
\newblock In {\em Proceedings of the 17th International Conference on
  Autonomous Agents and MultiAgent Systems}, pages 2040--2042, 2018.

\bibitem{peng2021phonetic}
Haiyun Peng, Yukun Ma, Soujanya Poria, Yang Li, and Erik Cambria.
\newblock Phonetic-enriched text representation for chinese sentiment analysis
  with reinforcement learning.
\newblock {\em Information Fusion}, 70:88--99, 2021.

\bibitem{pinto2017robust}
Lerrel Pinto, James Davidson, Rahul Sukthankar, and Abhinav Gupta.
\newblock Robust adversarial reinforcement learning.
\newblock In {\em International Conference on Machine Learning}, pages
  2817--2826. PMLR, 2017.

\bibitem{russo2019optimal}
Alessio Russo and Alexandre Proutiere.
\newblock Optimal attacks on reinforcement learning policies.
\newblock {\em arXiv preprint arXiv:1907.13548}, 2019.

\bibitem{sun2020stealthy}
Jianwen Sun, Tianwei Zhang, Xiaofei Xie, Lei Ma, Yan Zheng, Kangjie Chen, and
  Yang Liu.
\newblock Stealthy and efficient adversarial attacks against deep reinforcement
  learning.
\newblock In {\em Proceedings of the AAAI Conference on Artificial
  Intelligence}, volume~34, pages 5883--5891, 2020.

\bibitem{sutton2018reinforcement}
Richard~S Sutton and Andrew~G Barto.
\newblock {\em Reinforcement learning: An introduction}.
\newblock MIT press, 2018.

\bibitem{szegedy2014intriguing}
Christian Szegedy, Wojciech Zaremba, Ilya Sutskever, Joan Bruna, Dumitru Erhan,
  Ian Goodfellow, and Rob Fergus.
\newblock Intriguing properties of neural networks.
\newblock In {\em 2nd International Conference on Learning Representations,
  ICLR 2014}, 2014.

\bibitem{tretschk2018sequential}
Edgar Tretschk, Seong~Joon Oh, and Mario Fritz.
\newblock Sequential attacks on agents for long-term adversarial goals.
\newblock {\em arXiv preprint arXiv:1805.12487}, 2018.

\bibitem{wu2021adversarial}
Xian Wu, Wenbo Guo, Hua Wei, and Xinyu Xing.
\newblock Adversarial policy training against deep reinforcement learning.
\newblock In {\em 30th $\{$USENIX$\}$ Security Symposium ($\{$USENIX$\}$
  Security 21)}, 2021.

\bibitem{xiao2019characterizing}
Chaowei Xiao, Xinlei Pan, Warren He, Jian Peng, Mingjie Sun, Jinfeng Yi,
  Mingyan Liu, Bo~Li, and Dawn Song.
\newblock Characterizing attacks on deep reinforcement learning.
\newblock {\em arXiv preprint arXiv:1907.09470}, 2019.

\bibitem{zhang2021real}
Ke~Zhang, Yuanqing Li, Jingyu Wang, Erik Cambria, and Xuelong Li.
\newblock Real-time video emotion recognition based on reinforcement learning
  and domain knowledge.
\newblock {\em IEEE Transactions on Circuits and Systems for Video Technology},
  2021.

\bibitem{zhao2020blackbox}
Yiren Zhao, Ilia Shumailov, Han Cui, Xitong Gao, Robert Mullins, and Ross
  Anderson.
\newblock Blackbox attacks on reinforcement learning agents using approximated
  temporal information.
\newblock In {\em 2020 50th Annual IEEE/IFIP International Conference on
  Dependable Systems and Networks Workshops (DSN-W)}, pages 16--24. IEEE, 2020.

\end{thebibliography}

\end{document}